\documentclass[runningheads]{llncs}

\usepackage{graphicx}
\usepackage{multicol}
\usepackage{footmisc}
\usepackage{dsfont}
\usepackage{enumitem}
\usepackage{amsmath}
\usepackage{listings}
\usepackage{indentfirst}
\usepackage{multicol}
\usepackage{multirow}
\usepackage{graphicx}
\usepackage{ulem}
\usepackage{float}
\usepackage{array, makecell}
\usepackage{tabularx}
\usepackage{rotating}
\usepackage{hyperref}
\usepackage{booktabs}

\usepackage{color}

\newsavebox\CBox
\def\textBF#1{\sbox\CBox{#1}\resizebox{\wd\CBox}{\ht\CBox}{\textbf{#1}}}

\newcommand{\norm}[1]{\left\lVert#1\right\rVert}

\DeclareMathOperator{\sign}{sign}

\newcommand{\todov}[1]{{\color{black}#1}}

\usepackage[font=small,skip=2pt]{caption}
  
  \renewcommand{\textfloatsep}{1ex}
\title{Scalable and Interpretable One-class SVMs with Deep Learning and Random Fourier Features}
\author{Minh-Nghia Nguyen \and Ngo Anh Vien}
\institute{School of Electronics, Electrical Engineering and Computer Science, \\ Queen's University Belfast, UK, \\
\email{mnguyen04@qub.ac.uk, v.ngo@qub.ac.uk}}
\titlerunning{Scalable \& Interpretable OC-SVMs with Deep learning and Random features}

\begin{document}
\maketitle

\begin{abstract}
One-class support vector machine (OC-SVM) for a long time has been one of the most effective anomaly detection methods and extensively adopted in both research as well as industrial applications. The biggest issue for OC-SVM is yet the capability to operate with large and high-dimensional datasets due to optimization complexity. Those problems might be mitigated via dimensionality reduction techniques such as manifold learning or autoencoder. However, previous work often treats representation learning and anomaly prediction separately. In this paper, we propose autoencoder based one-class support vector machine (AE-1SVM) that brings OC-SVM, with the aid of random Fourier features to approximate the radial basis kernel, into deep learning context by combining it with a representation learning architecture and jointly exploit stochastic gradient descent to obtain end-to-end training. Interestingly, this also opens up the possible use of gradient-based attribution methods to explain the decision making for anomaly detection, which has ever been challenging as a result of the implicit mappings between the input space and the kernel space. To the best of our knowledge, this is the first work to study the interpretability of deep learning in anomaly detection. We evaluate our method on a wide range of unsupervised anomaly detection tasks in which our end-to-end training architecture achieves a performance significantly better than the previous work using separate training.
\end{abstract}

\section{Introduction}

Anomaly detection (AD), also known as outlier detection, is a unique class of machine learning that has a wide range of important applications, including intrusion detection in networks and control systems, fault detection in industrial manufacturing procedures, diagnosis of certain diseases in medical areas by identifying outlying patterns in medical images or other health records, cyber-security, etc. AD algorithms are identification processes that are able to single out items or events that are different from an expected pattern, or that have significantly lower frequencies compared to others in a dataset \cite{GrubbsOutliers,chandola2009anomaly}.

In the past, there has been substantial effort in using traditional machine learning techniques for both supervised and unsupervised AD such as principal component analysis (PCA) \cite{hotelling1933analysis,candes2011robust,ChalapathyMC17}, one-class support vector machine (OC-SVM) \cite{OCSVM,tax2004support,R1SVM}, isolation forests \cite{IsolationForest}, clustering based methods such as $k$-means, and Gaussian mixture model (GMM) \cite{barnett1974outliers,zimek2012survey,kim2012robust,xiong2011group}, etc. Notwithstanding, they often becomes inefficient when being used in high-dimensional problems because of high complexity and the absence of an integrated efficient dimensionality reduction approach. There is recently a growing interest in using deep learning techniques to tackle this issue. Nonetheless, most previous work still relies on two-staged or separate training in which a low-dimensional space is firstly learned via an autoencoder. For example, the work in \cite{HighDimSVM} simply proposes a hybrid architecture with a deep belief network to reduce the dimensionality of the input space and separately applies the learned feature space to a conventional OC-SVM. Robust deep autoencoder (RDA) \cite{RobustDeepAE} uses a structure that combines robust PCA and dimensionality reduction by autoencoder. However, this two-stage method is not able to learn efficient features for AD problems, especially when the dimensionality grows higher because of decoupled learning stages. More similar to our approach, deep clustering embedding (DEC) \cite{DEC} is a state-of-the-art algorithm that integrates unsupervised autoencoding network with clustering. Even though clustering is often considered as a possible solution to AD tasks, DEC is designed to jointly optimize the latent feature space and clustering, thus would learn a latent feature space that is more efficient to clustering rather than AD. 

End-to-end training of dimensionality reduction and AD has recently received much interest, such as the frameworks using deep energy-based model \cite{ZhaiCLZ16}, autoencoder combined with Gaussian mixture model \cite{zong2018deep}, generative adversarial networks (GAN) \cite{schlegl2017unsupervised,zenati2018efficient}. Nonetheless, these methods are based on density estimation techniques to detect anomalies as a by-product of unsupervised learning, therefore might not be efficient for AD. They might assign high density if there are many proximate anomalies (a new cluster or mixture might be established for them), resulting in false negative cases.



One-class support vector machine is one of the most popular techniques for unsupervised AD. OC-SVM is known to be insensitive to noise and outliers in the training data. 
Still, the performance of OC-SVM in general is susceptible to the dimensionality and complexity of the data \cite{SVMScaling}, while their training speed is also heavily affected by the size of the datasets. As a result, conventional OC-SVM may not be desirable in big data and high-dimensional AD applications. To tackle these issues, previous work has only performed dimensionality reduction via deep learning and OC-SVM based AD separately. Notwithstanding, separate dimensionality reduction might have a negative effect on the performance of the consequential AD, since important information useful for identifying outliers can be interpreted differently in the latent space. On the other hand, to the best of our knowledge, studies on the application of kernel approximation and stochastic gradient descent (SGD) on OC-SVM have been lacking: most of the existing works only apply random Fourier features (RFF) \cite{KernelApprox} to the input space and treat the problem as a linear support vector machine (SVM); meanwhile, \cite{Pegasos,SVMScaling} have showcased the prospect of using SGD to optimize SVM, but without the application of kernel approximation.

Another major issue in joint training with dimensionality reduction and AD is the interpretability of the trained models, that is, the capability to explain the reasoning for why they detect the samples as outliers, with respect to the input features. Very recently, explanation for black-box deep learning models has been brought about and attracted a respectable amount of attention from the machine learning research community. Especially, gradient-based explanation (attribution) methods \cite{ExplainClassification,SaliencyMaps,ancona2018towards} are widely studied as protocols to address this challenge. The aim of the approach is to analyse the contribution of each neuron in the input space of a neural network to the neurons in its latent space by calculating the corresponding gradients. As we will demonstrate, this same concept can be applied to kernel-approximated SVMs to score the importance of each input feature to the margin that separates the decision hyperplane.

Driven by those reasoning, in this paper we propose AE-1SVM that is an end-to-end autoencoder based OC-SVM model combining dimensionality reduction and OC-SVM for large-scale AD. RFFs are applied to approximate the RBF kernel, while the input of OC-SVM is fed directly from a deep autoencoder that shares the objective function with OC-SVM such that dimensionality reduction is forced to learn essential pattern assisting the anomaly detecting task. On top of that, we also extend gradient-based attribution methods on the proposed kernel-approximate OC-SVM as well as the whole end-to-end architecture to analyse the contribution of the input features on the decision making of the OC-SVM.

The remainder of the paper is organised as follows. Section \ref{Background} reviews the background on OC-SVM, kernel approximation, and gradient-based attribution methods. Section \ref{model_section} introduces the combined architecture that we have mentioned. In Section \ref{gradient_section}, we derive expressions and methods to obtain the end-to-end gradient of the OC-SVM's decision function with respect to the input features of the deep learning model. Experimental setups, results, and analyses are presented in Section \ref{exp_section}. Finally, Section \ref{conclusions} draws the conclusions for the paper.

\section{Background\label{Background}}
In this section, we briefly describe the preliminary background knowledge that is referred to in the rest of the paper.
\subsection{One-class support vector machine}
OC-SVM \cite{OCSVM} for unsupervised anomaly detection extends the idea of support vector method that is regularly applied in classification. While classic SVM aims to find the hyperplane to maximize the margin separating the data points, in OC-SVM the hyperplane is learned to best separate the data points from the origin. SVMs in general have the ability to capture non-linearity thanks to the use of kernels. The kernel method maps the data points from the input feature space in $\mathcal{R}^d$ to a higher-dimensional space in $\mathcal{R}^D$ (where $D$ is potentially infinite), where the data is linearly separable, by a transformation $\mathcal{R}^d \to \mathcal{R}^D$. The most commonly used kernel is the radial basis function (RBF) kernel defined by a similarity mapping between any two points $x$ and $x'$ in the input feature space, formulated by $ K(x,x') = \exp(-\frac{\norm{x-x'}^2}{2\sigma^2})$, with $\sigma$ being a kernel bandwidth.

Let $\mathsf{w}$ and $\rho$ denote the vectors indicating the weights of all dimensions in the kernel space and the offset parameter determining the distance from the origin to the hyperplane, respectively. The objective of OC-SVM is to separate all data points from the origin by a maximum margin with respect to some constraint relaxation, and is written as a quadratic program as follows:

\begin{align}
\min_{\mathsf{w}, \xi, \rho} \frac{1}{2} \norm{\mathsf{w}}^2 - \rho + \frac{1}{\nu n}\sum_{i=1}^{n}\xi_i, \\
\text{subject to } \mathsf{w}^T \phi(x_i) \geq \rho - \xi_i, \xi_i \geq 0 \notag.
\label{ocsvm_obj}
\end{align}
where $\xi_i$ is a slack variable and $\nu$ is the regularization parameter. Theoretically, $\nu$ is the upper bound of the fraction of anomalies in the data, and also the main tuning parameter for OC-SVM. Additionally, by replacing $\xi_i$ with the hinge loss, we have the unconstrained objective function as

\begin{align}
\min_{\mathsf{w}, \rho} \frac{1}{2} \norm{\mathsf{w}}^2 - \rho + \frac{1}{\nu n}\sum_{i=1}^{n}\max(0, \rho - \mathsf{w}^T \phi(x_i)).
\end{align}

Let $g(x) = \mathsf{w}.\phi(x_i) - \rho$, the decision function of OC-SVM is
\begin{align}
f(x) = \sign(g(x)) = \begin{cases} 1 & \mbox{if } g(x) \geq 0 \\ -1 & \mbox{if } g(x) < 0 \end{cases}.
\end{align}

The optimization problem of SVM in \eqref{ocsvm_obj} is usually solved as a convex optimization problem in the dual space with the use of Lagrangian multipliers to reduce complexity while increasing solving feasibility. LIBSVM \cite{LIBSVM} is the most popular library that provides efficient optimization algorithms to train SVMs, and has been widely adopted in the research community. Nevertheless, solving SVMs in the dual space can be susceptible to the data size, since the function $K$ between each pair of points in the dataset has to be calculated and stored in a matrix, resulting in an $O(n^2)$ complexity, where $n$ is the size of the dataset.

\subsection{Kernel approximation with random Fourier features}
To address the scalability problem of kernel machines, approximation algorithms have been introduced and widely applied, with the most two dominant being Nystr\"{o}em \cite{Nystroem} and random Fourier features (RFF) \cite{KernelApprox}. In this paper, we focus on RFF since it has lower complexity and does not require pre-training. The method is based on the Fourier transform of the kernel function, given by a Gaussian distribution:
\begin{align}
p(\omega) = \mathcal{N}(0, \sigma^{-2}\mathds{I})
\end{align}
where $\mathds{I}$ is the identity matrix and $\sigma$ is an adjustable parameter representing the standard deviation of the Gaussian process.

From the distribution $p$, $D$ independent and identically distributed weights $\omega_1, \omega_2, ..., \omega_D$ are drawn. In the original work \cite{KernelApprox}, two mappings are introduced, which are:

\begin{itemize}
\item The combined $cosine$ and $sine$ mapping as $z_\omega(x) = \begin{bmatrix} cos(\omega^T x) & sin(\omega^T x) \end{bmatrix} ^T$, which leads to the complete mapping being defined as follows:
\begin{align}
z(x) = \sqrt{\frac{1}{D}}\begin{bmatrix} cos(\omega_1^T x) & ... & cos(\omega_D^T x) & sin(\omega_1^T x) & ... & sin(\omega_D^T x) \end{bmatrix} ^T,
\label{RFF_mappings}
\end{align}
\item The offset $cosine$ mapping as $z_\omega(x) = \sqrt{2} cos(\omega^T x + b)$,
where the offset parameter $b \sim \mathit{U}(0, 2\pi)$. Consequently, the complete mapping in this case is
\begin{align}
z(x) = \sqrt{\frac{2}{D}}\begin{bmatrix} cos(\omega_1^T x + b) & ... & cos(\omega_D^T x + b)\end{bmatrix} ^T.
\end{align}
\end{itemize}

It has been proven in \cite{RFFError} that the former mapping outperforms the latter one in approximating RBF kernels due to the fact that no phase shift is introduced as a result of the offset variable. Therefore, in this paper, we only consider the combined $sine$ and $cosine$ mapping.

Applying the kernel approximation mappings to \eqref{ocsvm_obj}, the unconstrained OC-SVM objective function with hinge loss becomes
\begin{align}
\min_{\mathsf{w}, \rho} \frac{1}{2} \norm{\mathsf{w}}^2 - \rho + \frac{1}{\nu n}\sum_{i=1}^{n}\max(0, \rho - \mathsf{w}^T z(x_i)),
\end{align}
which is equivalent to a OC-SVM in the approximated kernel space $\mathcal{R}^D$, and thus the optimization problem is more trivial, despite the dimensionality of $\mathcal{R}^D$ being higher than that of $\mathcal{R}^d$.

\subsection{Gradient-based explanation methods}
Gradient-based methods exploit the gradient of the latent nodes in a neural network with respect to the input features to rate the attribution of each input to the output of the network. In the recent years, many research studies \cite{SaliencyMaps,AxiomaticAttribution,ExplainNonLinearClassification,ancona2018towards} have applied this approach to explain the classification decision and sensitivity of input features in deep neural networks and especially convolutional neural networks. Intuitively, an input dimension $\mathbf{x}_i$ has larger contribution to a latent node $\mathbf{y}$ if the gradient of $\mathbf{y}$ with respect to $\mathbf{x}_i$ is higher, and vice versa. 

Instead of using purely gradient as a quantitative factor, various extensions of the method has been developed, including Gradient*Input \cite{GradientxInput}, Integrated gradients \cite{AxiomaticAttribution}, or DeepLIFT \cite{DeepLIFT}. The most recent work \cite{ancona2018towards} showed that these methods are strongly related and proved conditions of equivalence or approximation between them. In addition, other non gradient-based can be re-formulated to be implemented easily like gradient-based.


\section{Deep autoencoding one-class SVM \label{model_section}}
\begin{figure}[H]
\centering
\includegraphics[trim=0cm 0cm 0cm 1.8cm, width=0.576\linewidth]{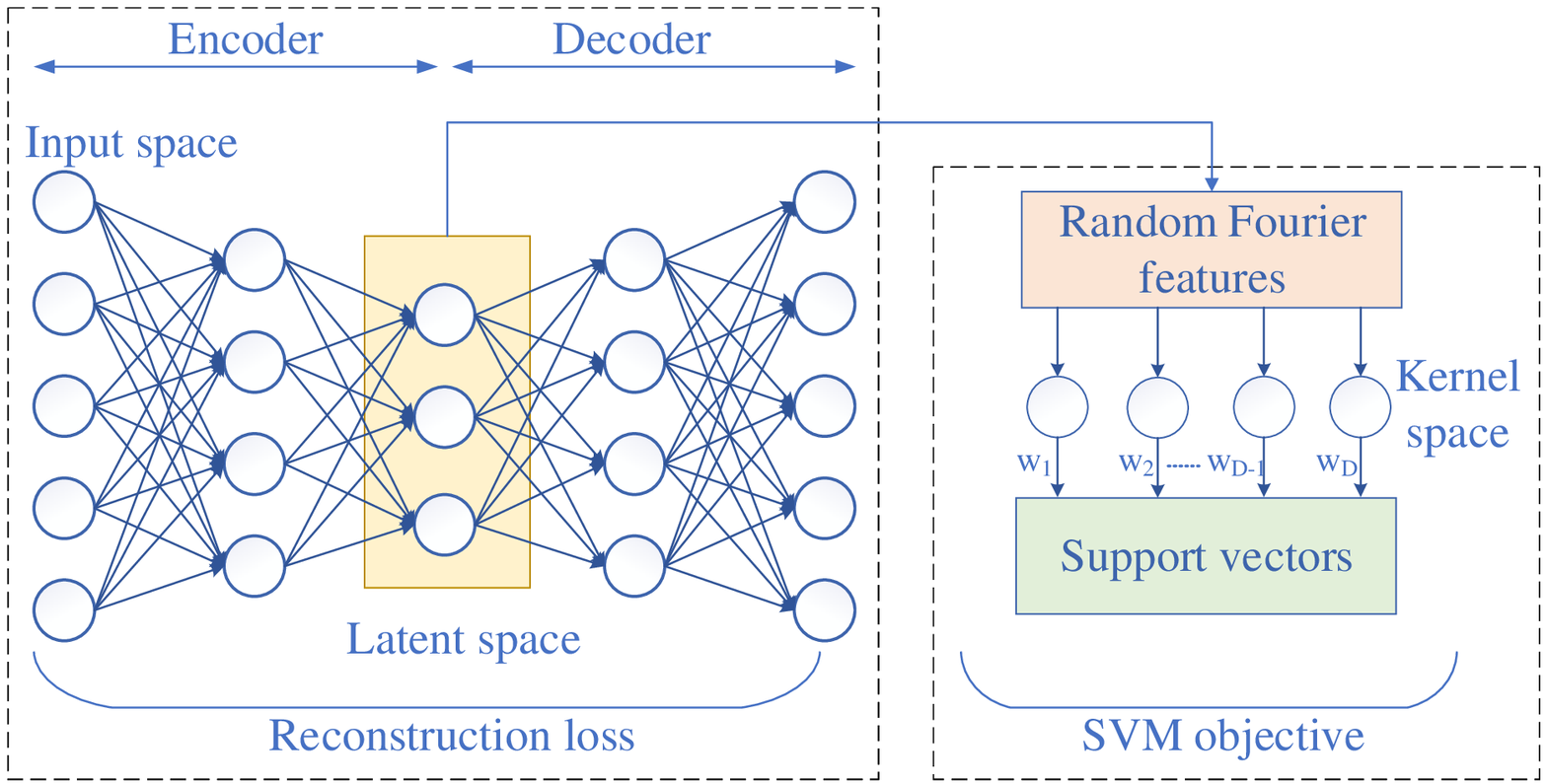} 
\quad \quad 
\includegraphics[trim=0cm 0cm 0cm 1.8cm, width=0.35\linewidth]{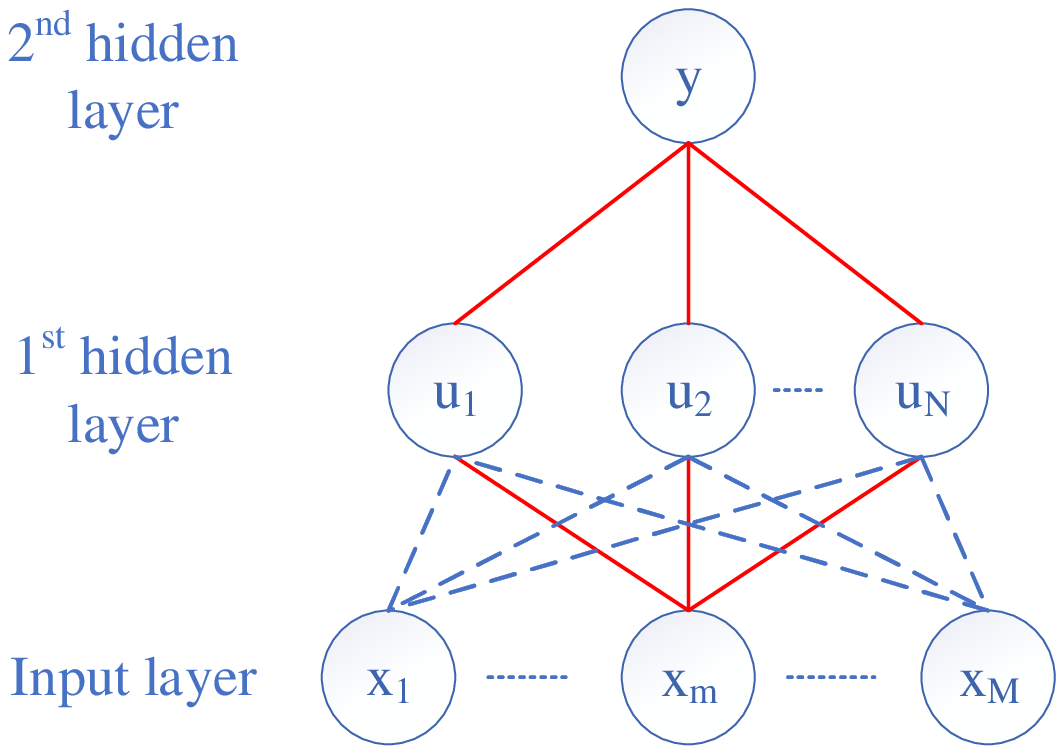}
\caption{(Left) Illustration of the Deep autoencoding One-class SVM architecture. (Right) Connections between input layer and hidden layers of a neural network \label{DeepAEOCSVM_model}}
\end{figure}
In this section, we present our combined model, namely Deep autoencoding One-class SVM (AE-1SVM), based on OC-SVM for anomaly detecting tasks in high-dimensional and big datasets. The model consists of two main components, as illustrated in Figure \ref{DeepAEOCSVM_model} (Left): 
\begin{itemize}
\item A deep autoencoder network for dimensionality reduction and feature representation of the input space. 
\item An OC-SVM for anomaly prediction based on support vectors and margin. The RBF kernel is approximated using random Fourier features.
\end{itemize} 

The bottleneck layer of the deep autoencoder network is forwarded directly into the Random features mapper as the input of the OC-SVM. By doing this, the autoencoder network is pressed to optimize its variables to represent the input features in the direction that supports the OC-SVM in separating the anomalies from the normal class. 

\todov{Let us denote $\mathbf{x}$ as the input of the deep autoencoder, $\mathbf{x}'$ as the reconstructed value of $\mathbf{x}$, and $x$ as the latent space of the autoencoder}. In addition, $\theta$ is the set of parameters of the autoencoder. As such, the joint objective function of the model regarding the autoencoder parameters, the OC-SVM's weights, and its offset is as follows:
\begin{align}
Q(\theta,\mathsf{w}, \rho)  = \alpha L(\mathbf{x}, \mathbf{x}') + \frac{1}{2} \norm{\mathsf{w}}^2 - \rho + \frac{1}{\nu n}\sum_{i=1}^{n}\max(0, \rho - \mathsf{w}^T z(x_i))
\label{obj_func}
\end{align}

The components and parameters in \eqref{obj_func} are described below
\begin{itemize}
\item $L(\mathbf{x}, \mathbf{x}')$ is the reconstruction loss of the autoencoder, which is normally chosen to be the L2-norm loss $L(\mathbf{x}, \mathbf{x}') = \norm{\mathbf{x} - \mathbf{x}'}_2^2$.
\item Since SGD is applied, the variable $n$, which is formerly the number of training samples, becomes the batch size since the hinge loss is calculated using the data points in the batch. 
\item $z$ is the random Fourier mappings as defined in \eqref{RFF_mappings}. Due to the random features being data-independent, the standard deviation $\sigma$ of the Gaussian distribution has to be fine-tuned correlatively with the parameter $\nu$.
\item $\alpha$ is a hyperparameter controlling the trade-off between feature compression and SVM margin optimization.
\end{itemize} 

Overall, the objective function is optimized in conjunction using SGD with backpropagation. Furthermore, the autoencoder network can also be extended to a convolutional autoencoder, which is showcased in the experiment section.

\section{Interpretable autoencoding one-class SVM \label{gradient_section}}
In this section, we outline the method for interpreting the results of AE-1SVM using gradients and present illustrative example to verify its validity.

\subsection{Derivations of end-to-end gradients}
Considering an input $x$ of an RFF kernel-approximated OC-SVM with dimensionality $R^d$. In our model, $x$ is the bottleneck representation of the latent space in the deep autoencoder. The expression of the margin $g(x)$ with respect to the input $x$ is as follows:
\begin{align}
g(x) & = \sum_{j=1}^{D}\mathsf{w}_j z_{\omega_j}(x) - \rho \nonumber  = \sqrt{\frac{1}{D}}\sum_{j=1}^{D}\big[\mathsf{w}_jcos(\omega_j^T x) + \mathsf{w}_{D+j}sin(\omega_j^T x)\big] - \rho \nonumber \\
& = \sqrt{\frac{1}{D}}\sum_{j=1}^{D}\big[\mathsf{w}_jcos(\sum_{k=1}^{d}\omega_{jk}x_k) + \mathsf{w}_{D+j}sin(\sum_{k=1}^{d}\omega_{jk}x_k)\big] - \rho .
\end{align}
As a result, the gradient of the margin function on each input dimension $k = 1, 2, ..., d$ can be calculated as
\begin{align}
\frac{\partial g}{\partial x_k} & = \sqrt{\frac{1}{D}}\sum_{j=1}^{D}\omega_{jk}\big[-\mathsf{w}_j sin(\sum_{k=1}^{d}\omega_{jk}x_k) +\mathsf{w}_{j+D} cos(\sum_{k=1}^{d}\omega_{jk}x_k)\big].
\label{gradient_svm}
\end{align}

Next, we can derive the gradient of the latent space nodes with respect to the deep autoencoder's input layer (extension to convolutional autoencoder is straightforward). In general, considering a neural network with $M$ input neurons $x_m, m = 1, 2, ..., M$, and the first hidden layer having $N$ neurons $u_n, n = 1, 2, ..., N$, as depicted in Figure \ref{DeepAEOCSVM_model} (Right). The gradient of $u_n$ with respect to $x_m$ can be derived as

\begin{align}
G(x_m, u_n) = \frac{\partial u_n}{\partial x_m} = w_{mn}\sigma '(x_m w_{mn} + b_{mn})\sigma(x_m w_{mn} + b_{mn}),
\label{gradient_layer1}
\end{align}


where $\sigma(x_m w_{mn} + b_{mn}) = u_n$, $\sigma(.)$ is the activation function, $w_{mn}$ and $b_{mn}$ are the weight and bias connecting $x_m$ and $u_n$. The derivative of $\sigma$ is different for each activation function. For instance, with a sigmoid activation $\sigma$, the gradient $G(x_m, u_n)$ is computed as $\displaystyle w_{mn}u_n(1-u_n)$, while $G(x_m, u_n)$ is $w_{mn}(1-u_n^2)$ for $tanh$ activation function.

To calculate the gradient of neuron $y_l$ in the second hidden layer with respect to $x_m$, we simply apply the chain rule and sum rule as follows:
\begin{align}
G(x_m, y_l) = \frac{\partial y_l}{\partial x_m} & =  \sum_{n=1}^{N}\frac{\partial y_l}{\partial u_n}\frac{\partial u_n}{\partial x_m} = \sum_{n=1}^{N}G(u_n, y_l)G(x_m, u_n).
\label{gradient_layer2}
\end{align}

The gradient $G(u_n, y_l)$ can be obtained in a similar manner to \eqref{gradient_layer1}. By maintaining the values of $G$ at each hidden layer, the gradient of any hidden or output layer with respect to the input layer can be calculated. Finally, combining this and \eqref{gradient_svm}, we can get the end-to-end gradient of the OC-SVM margin with respect to all input features. Besides, state-of-the-art machine learning frameworks like TensorFlow \todov{ also implements automatic differentiation \cite{tensorflow2015-whitepaper} that simplifies the procedures for computing those gradient values}.

Using the obtained values, the decision making of the AD model can be interpreted as follows:

\begin{itemize}
\item For an outlying sample, the dimension which has higher gradient indicates a higher contribution to the decision making of the ML model. In other words, the sample is further to the boundary in that particular dimension.

\item For each mentioned dimension, if the gradient is positive, the value of the feature in that dimension is lesser than the the lower limit of the boundary. In contrast, if the gradient holds a negative value, the feature exceeds the level of the normal class.
\end{itemize}

\subsection{Illustrative example}

\begin{figure}
\centering
\includegraphics[trim=0cm 0cm 0cm 1.5cm, width=0.8\linewidth]{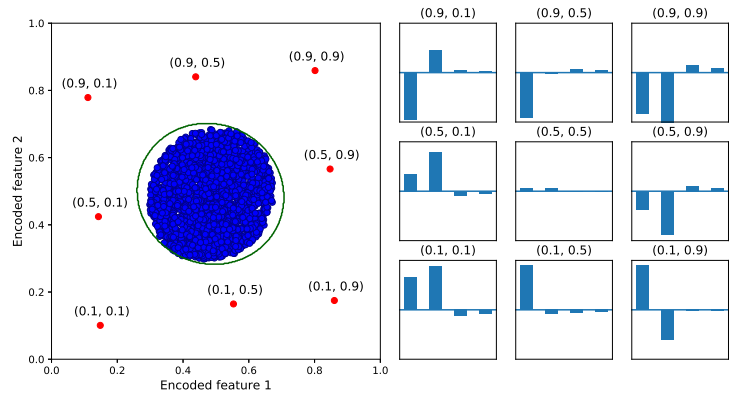}
\caption{An illustrative example of gradient-based explanation methods. The left figure depicts the encoded 2-dimensional feature space from a 4-dimension dataset. The nine graphs on the right plot the gradient of the margin function with respect to the four original features for each testing point. Only the coordinates of first two dimensions are annotated. \label{gradient_explain}}
\end{figure}

Figure \ref{gradient_explain} presents an illustrative example of interpreting anomaly detecting results using gradients. We generate 1950 four-dimensional samples as normal instances, where the first two features are uniformly generated such that they are inside a circle with center $C(0.5, 0.5)$. The third and fourth dimensions are drawn uniformly in the range $[-0.2, 0.2]$ so that the contribution of them are significantly less than the other two dimensions. In contrast, 50 anomalies are created which have the first two dimensions being far from the mentioned circle, while the last two dimensions has a higher range of $[-2, 2]$. The whole dataset including both the normal and anomalous classes are trained with the proposed AE-1SVM model with a bottleneck layer of size 2 and sigmoid activation.

The figure on the left shows the representation of the 4D dataset on a 2-dimensional space. Expectedly, it captures most of the variability from only the first two dimensions. Furthermore, we plot the gradients of 9 different anomalous samples, with the two latter dimensions being randomized, and overall, the results have proven the aforementioned interpreting rules. It can easily be observed that the contribution of the third and fourth dimensions to the decision making of the model is always negligible. Among the first two dimensions, the ones having the value of 0.1 or 0.9 has the corresponding gradients perceptibly higher than those being 0.5, as they are further from the boundary and the sample can be considered "more anomalous" in that dimension. Besides, the gradient of the input 0.1 is always positive due to the fact that it is lower than the normal level. In contrast, the gradient of the input 0.9 is consistently negative.

\section{Experimental results \label{exp_section}}
We present qualitative empirical analysis to justify the effectiveness of the AE-1SVM model in terms of accuracy and improved training/testing time. The objective is to compare the proposed model with conventional and state-of-the-art AD methods over synthetic and well-known real world data \footnote{All code for reproducibility is available at https://github.com/minh-nghia/AE-1SVM}.

\subsection{Datasets}
We conduct experiments on one generated datasets and five real-world datasets (we assume all tasks are unsupervised AD) as listed below in Table \ref{datasets}. The descriptions of each individual dataset is as follows:

\begin{itemize}
\item \textbf{Gaussian}: This dataset is taken into account to showcase the performance of the methods on high-dimensional and large data. The normal samples are drawn from a normal distribution with zero mean and standard deviation $\sigma=1$, while $\sigma=5$ for the anomalous instances. Theoretically, since the two groups have different distributional attributes, the AD model should be able to separate them.

\item \textbf{ForestCover}: From the ForestCover/Covertype dataset \cite{UCI}, the class 2 is extracted as the normal class, and class 4 is chosen as the anomaly class.

\item \textbf{Shuttle}: From the Shuttle dataset \cite{UCI}, we select the normal samples from classes 2, 3, 5, 6, 7, while the outlier group is made of class 1.

\item \textbf{KDDCup99}: The popular KDDCup99 dataset \cite{UCI} has approximately 80\% proportion as anomalies. Therefore, from the 10-percent subset, we randomly select 5120 samples from the outlier classes to form the anomaly set such that the contamination ratio is 5\%. The categorical features are extracted using one-hot encoding, making 118 features in the raw input space. 

\item \textbf{USPS}: We select from the U.S Postal Service handwritten digits dataset \cite{USPS} 950 samples from digit 1 as normal data, and 50 samples from digit 7 as anomalous data, as the appearance of the two digits are similar. The size of each image is 16 $\times$ 16, resulting in each sample being a flatten vector of 256 features.

\item \textbf{MNIST}: From the MNIST dataset \cite{mnist}, 5842 samples of digit '4' are chosen as normal class. On the other hand, the set of outliers contains 100 digits from classes '0', '7', and '9'. This task is challenging due to the fact that many digits '9' are remarkably similar to digit '4'. Each input sample is a flatten vector with 784 dimensions.

\end{itemize}

\begin{table}[h]
\centering
\begin{tabularx}{\linewidth}{ X X X X }
\hline
Dataset & Dimensions & Normal instances & Anomalies rate (\%) \\
\hline
\midrule
Gaussian & 512 & 950 & 5.0 \\
\hline
ForestCover & 54 & 581012 & 0.9 \\
\hline
Shuttle & 9 & 49097 & 7.2 \\
\hline
KDDCup99 & 118 & 97278 & 5.0  \\
\hline
USPS & 256 & 950 & 5.0  \\
\hline
MNIST & 784 & 5842 & 1.7  \\
\hline
\end{tabularx}
\setlength{\abovecaptionskip}{0.2cm}
\caption{Summary of the datasets used for comparison in the experiments. \label{datasets}}
\end{table}

\subsection{Baseline methods}
Variants of OC-SVM and several state-of-the-art methods are selected as baselines to compare the performance with the AE-1SVM model. Different modifications of the conventional OC-SVM are considered. First, we take into account the version where OC-SVM \todov{with RBF kernel} is trained directly on the raw input. Additionally, to give more impartial justifications, a version where an autoencoding network exactly identical to that of the AE-1SVM model is considered. We use the same number of training epochs to AE-1SVM to investigate the ability of AE-1SVM to force the dimensionality reduction network to learn better representation of the data. The OC-SVM is then trained on the encoded feature space, and this variant is also similar to the approach given in \cite{HighDimSVM}.

The following methods are also considered as baselines to examine the anomaly detecting performance of the proposed model:
\begin{itemize}

\item \textbf{Isolation Forest} \cite{IsolationForest}: This ensemble method revolves around the idea that the anomalies in the data have significantly lower frequencies and are different from the normal points.
\item \textbf{Robust Deep Autoencoder (RDA)} \cite{RobustDeepAE}: In this algorithm, a deep autoencoder is constructed and trained such that it can decompose the data into two components. The first component contains the latent space representation of the input, while the second one is comprised of the noise and outliers that are difficult to reconstruct.

\item \textbf{Deep Clustering Embeddings (DEC)} \cite{DEC}: This algorithm combines unsupervised autoencoding network with clustering. As outliers often locate in sparser clusters or are far from their centroids, we apply this method into AD and calculate the anomaly score of each sample as a product of its distance to the centroid and the density of the cluster it belongs to.
\end{itemize}
\subsection{Evaluation metrics}
In all experiments, the area under receiver operating characteristic (AUROC) and area under the Precesion-Recall curve (AUPRC) are applied as metrics to evaluate and compare the performance of AD methods. Having a high AUROC is necessary for a competent model, whereas AUPRC often highlights the difference between the methods regarding imbalance datasets \cite{PRROC}. The testing procedure follows the unsupervised setup, where each dataset is split with 1:1 ratio, and the entire training set including the anomalies is used for training the model. The output of the models on the test set is measured against the ground truth using the mentioned scoring metrics, with the average scores and approximal training and testing time of each algorithm after 20 runs being reported.

\subsection{Model configurations}
In all experiments, we employ the sigmoid activation function and implement the architecture using TensorFlow \cite{tensorflow2015-whitepaper}. The initial weights of the autoencoding networks are generated according to Xavier's method \cite{xavierInit}. The optimizing algorithm of choice is Adam \cite{AdamOptimizer}. We also discover that for the random Fourier features, a standard deviation $\sigma = 3.0$ produces satisfactory results for all datasets. For other parameters, the network configurations of AE-1SVM for each individual dataset are as in Table \ref{configs} below.

\begin{table}[h]
\centering
\begin{tabularx}{\linewidth}{ l l c c X X X}
\hline
Dataset & Encoding layers & $\nu$ & $\alpha$ & RFF & Batch size & Learning rate\\
\hline
\hline
\textBF{Gaussian} & \{128, 32\} & 0.40 & 1000 & 500 & 32 & 0.01 \\
\hline
\textBF{ForestCover} & \{32, 16\} & 0.30 & 1000 & 200 & 1024 & 0.01 \\
\hline
\textBF{Shuttle} & \{6, 2\} & 0.40 & 1000 & 50 & 16 & 0.001 \\
\hline
\textBF{KDDCup99} & \{80, 40, 20\} & 0.30 & 10000 & 400 & 128 & 0.001 \\
\hline 
\textBF{USPS} & \{128, 64, 32\} & 0.28 & 1000 & 500 & 16 & 0.005 \\
\hline 
\textBF{MNIST} & \{256, 128\} & 0.40 & 1000 & 1000 & 32 & 0.001 \\
\hline 

\end{tabularx}
\setlength{\abovecaptionskip}{0.2cm}
\caption{Summary of network configurations and training parameters of AE-1SVM used in the experiments. \label{configs}}
\end{table}

For the MNIST dataset, we additionally implement a convolutional autoencoder with pooling and unpooling layers: conv1($5\times5\times 16$), pool1($2 \times 2$), conv2($5\times5\times 9$), pool2($2 \times 2$) and a feed-forward layer afterward to continue compressing into 49 dimensions; the decoder: a feed-forward layer afterward of $49\times9$ dimensions, then deconv1($5\times5\times 9$), unpool1($2 \times 2$), deconv2($5\times5\times 16$), unpool2($2 \times 2$), then a feed-forward layer of 784 dimensions. The dropout rate is set to 0.5 in this convolutional autoencoder network. 

For each baseline methods, the best set of parameters are selected. In particular, for different variants of OC-SVM, the optimal \todov{values for parameter $\nu$ and the RBF kernel width are} exhaustively searched. Likewise, for Isolation forest, the fraction ratio is tuned around the anomalies rate for each dataset. For RDA, DEC, as well as OC-SVM variants that involves auto-encoding network for dimensionality reduction, the autoencoder structures exactly identical to AE-1SVM are used, while the $\lambda$ hyperparameter in RDA is also adjusted as it is the most important factor of the algorithm. 

\subsection{Results}

\begin{figure}
\centering
\includegraphics[width=0.8\linewidth]{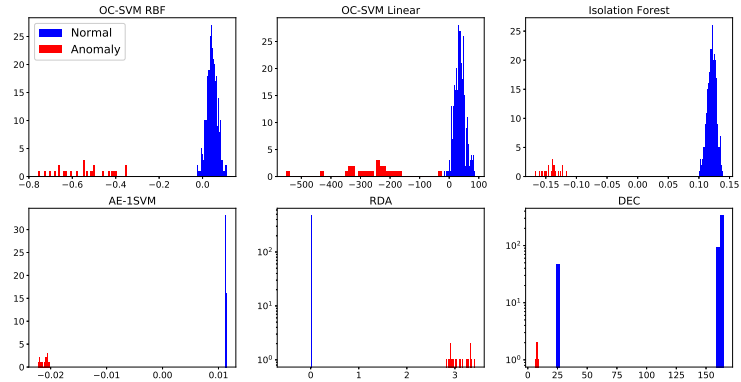}
\caption{Histograms of decision scores of AE-1SVM and other baseline methods. \label{histogram}}
\end{figure}

Firstly, for the Gaussian dataset, the histograms of the decision scores obtained by different methods are presented in Figure \ref{histogram}. It can clearly be seen that AE-1SVM is able to single out all anomalous samples, while giving the best separation between the two classes.

For other datasets, the comprehensive results are given in Table \ref{results}. 
It is obvious that AE-1SVM outperforms conventional OC-SVM \todov{as well as the two-staged structure with decoupled autoencoder and OC-SVM} in terms of accuracy in all scenarios, and is always among the top performers. \todov{As we restrict the number of training epochs for the detached autoencoder to be same as that for AE-1SVM, its performance declines significantly and in some cases its representation is even worse than the raw input. This proves that AE-1SVM can attain more efficient features to support AD task given the similar time.} 

Other observations can also be made from the results. For ForestCover, only the AUROC score of Isolation Forest is close, but the AUPRC is significantly lower, with three time less than that of AE-1SVM, suggesting that it has to compensate a higher false alarm rate to identify anomalies correctly. Similarly, Isolation Forest slightly surpasses AE-1SVM in AUROC for Shuttle dataset, but is subpar in terms of AUPRC, thus can be considered less optimal choice.
autoencoder network in image processing contexts.  Analogous patterns can as well be noticed for other datasets. Especially, for MNIST, it is shown that the proposed method AE-1SVM can also operate under a convolutional autoencoder network in image processing contexts.

\begin{table}[H]
\centering
\begin{tabularx}{\columnwidth}{ X X X X X X X}
\toprule
\textBF{Dataset} & \multicolumn{2}{>{\hsize=2\hsize}X}{\textBF{Method}} & \textBF{AUROC} & \textBF{AUPRC} & \textBF{Train} & \textBF{Test} \\
\midrule
\multirow{6}{0.1pt}{\textBF{Forest} \textBF{Cover}} & \multicolumn{2}{>{\hsize=2.5\hsize}X}{OC-SVM raw input} & 0.9295 & 0.0553 & $6 \times 10^2$  & $2 \times 10^2$ \\ 
& \multicolumn{2}{>{\hsize=2.5\hsize}X}{OC-SVM encoded} & 0.7895 & 0.0689 & $2.5 \times 10^2$ & $8 \times 10^1$ \\
 & \multicolumn{2}{>{\hsize=2.5\hsize}X}{Isolation Forest} & 0.9396 & 0.0705 & $3 \times 10^1$  & $1 \times 10^1$  \\
& \multicolumn{2}{>{\hsize=2.5\hsize}X}{RDA} & 0.8683 & 0.0353 & $1 \times 10^2$ & $2 \times 10^0$ \\
& \multicolumn{2}{>{\hsize=2.5\hsize}X}{DEC} & 0.9181 & 0.0421 & $\mathbf{2 \times 10^1}$  & $4 \times 10^0$ \\
& \multicolumn{2}{>{\hsize=2.5\hsize}X}{AE-1SVM} & \textBF{0.9485} &  \textBF{0.1976} & $\mathbf{2 \times 10^1}$ & $\mathbf{7 \times 10^{-1}}$ \\

\hline
\midrule
\multirow{6}{0.1pt}{\textBF{Shuttle}} & \multicolumn{2}{>{\hsize=2.5\hsize}X}{OC-SVM raw input} & 0.9338 & 0.4383 & $2 \times 10^1$  & $5 \times 10^1$ \\ 
& \multicolumn{2}{>{\hsize=2.5\hsize}X}{OC-SVM encoded} & 0.8501 & 0.4151 & $2 \times 10^1$ & $2.5 \times 10^0$ \\

& \multicolumn{2}{>{\hsize=2.5\hsize}X}{Isolation Forest} & \textBF{0.9816} & 0.7694 & $2.5 \times 10^1$  & $1.5 \times 10^1$  \\
& \multicolumn{2}{>{\hsize=2.5\hsize}X}{RDA} & 0.8306 & 0.1872 & $3 \times 10^2$ & $2 \times 10^{-1}$ \\
& \multicolumn{2}{>{\hsize=2.5\hsize}X}{DEC} & 0.9010 & 0.3184 & $\mathbf{6 \times 10^0}$ & $1 \times 10^0$ \\
& \multicolumn{2}{>{\hsize=2.5\hsize}X}{AE-1SVM} & 0.9747 &  \textBF{0.9483} & $1 \times 10^1$ & $\mathbf{1 \times 10^{-1}}$ \\

\hline
\midrule
\multirow{7}{0.1pt}{\textBF{KDDCup}} & \multicolumn{2}{>{\hsize=2.5\hsize}X}{OC-SVM raw input} & 0.8881 & 0.3400 & $6 \times 10^1$  & $2 \times 10^1$ \\ 
& \multicolumn{2}{>{\hsize=2.5\hsize}X}{OC-SVM encoded} & 0.9518 & 0.3876 & $5 \times 10^1$  & $1 \times 10^1$ \\
 & \multicolumn{2}{>{\hsize=2.5\hsize}X}{Isolation Forest} & 0.9572 & 0.4148 & $2 \times 10^1$  & $5 \times 10^0$  \\
& \multicolumn{2}{>{\hsize=2.5\hsize}X}{RDA} & 0.6320 & 0.4347 & $1 \times 10^2$ & $5 \times 10^{-1}$ \\
& \multicolumn{2}{>{\hsize=2.5\hsize}X}{DEC} & 0.9496 & 0.3688 & $\mathbf{1 \times 10^1}$  & $2 \times 10^0$ \\
& \multicolumn{2}{>{\hsize=2.5\hsize}X}{AE-1SVM} & \textBF{0.9663} &  \textBF{0.5115} & $3 \times 10^1$ & $\mathbf{4.5 \times 10^{-1}}$ \\
& \multicolumn{2}{>{\hsize=2.5\hsize}X}{AE-1SVM (Full dataset)} & \textBF{0.9701} &  \textBF{0.4793} & $2 \times 10^2$ & $4 \times 10^{0}$ \\

\hline
\midrule
\multirow{6}{0.1pt}{\textBF{USPS}} & \multicolumn{2}{>{\hsize=2.5\hsize}X}{OC-SVM raw input} & 0.9747 & 0.5102 & $\mathbf{2 \times 10^{-2}}$  & $1.5 \times 10^{-2}$ \\ 
& \multicolumn{2}{>{\hsize=2.5\hsize}X}{OC-SVM encoded} & 0.9536 & 0.4722 & $6 \times 10^{0}$  & $4 \times 10^{-3}$ \\
 & \multicolumn{2}{>{\hsize=2.5\hsize}X}{Isolation Forest} & 0.9863 & 0.6250 & $2.5 \times 10^{-1}$  & $6 \times 10^{-2}$  \\
& \multicolumn{2}{>{\hsize=2.5\hsize}X}{RDA} & 0.9799 & 0.5681 & $1.5 \times 10^0$ & $1.5 \times 10^{-2}$ \\
& \multicolumn{2}{>{\hsize=2.5\hsize}X}{DEC} & 0.9263 & 0.7506 & $4 \times 10^0$  & $2.5 \times 10^{-2}$ \\
& \multicolumn{2}{>{\hsize=2.5\hsize}X}{AE-1SVM} & \textBF{0.9926} &  \textBF{0.8024} & $1 \times 10^1$ & $\mathbf{5 \times 10^{-3}}$ \\

\hline
\midrule
\multirow{7}{0.1pt}{\textBF{MNIST}} & \multicolumn{2}{>{\hsize=2.5\hsize}X}{OC-SVM raw input} & 0.8302 & 0.0819 & $\mathbf{2 \times 10^0}$  & $1 \times 10^0$ \\ 
& \multicolumn{2}{>{\hsize=2.5\hsize}X}{OC-SVM encoded} & 0.7956 & 0.0584 & $1 \times 10^{2}$ & $1 \times 10^{-1}$ \\

& \multicolumn{2}{>{\hsize=2.5\hsize}X}{Isolation Forest} & 0.7574 & 0.0533 & $4.5 \times 10^0$  & $1.5 \times 10^0$  \\
& \multicolumn{2}{>{\hsize=2.5\hsize}X}{RDA} & 0.8464 & 0.0855 & $1 \times 10^2$ & $2.5 \times 10^{-1}$ \\
& \multicolumn{2}{>{\hsize=2.5\hsize}X}{DEC} & 0.5522 & 0.0289 & $3.5 \times 10^1$ & $\mathbf{1.5 \times 10^{-1}}$ \\
& \multicolumn{2}{>{\hsize=2.5\hsize}X}{AE-1SVM} & 0.8119 &  0.0864 & $1.5 \times 10^2$ & $7 \times 10^{-1}$ \\
& \multicolumn{2}{>{\hsize=2.5\hsize}X}{CAE-1SVM} & \textBF{0.8564} &  \textBF{0.0885} & $3.5 \times 10^3$ & $1.5 \times 10^{0}$ \\

\bottomrule
\end{tabularx}
\setlength{\abovecaptionskip}{0.2cm}
\caption{Average AUROC, AUPRC, approximal train time and test time of the baseline methods and proposed method. Best results are displayed in boldface. \label{results}}
\end{table}

Regarding training time, AE-1SVM outperforms other methods for ForestCover, which is the largest dataset. For other datasets that have high sample size, namely KDDCup99 and Shuttle, it is still one of the fastest candidates. Furthermore, we also extend the KDDCup99 experiment and train AE-1SVM model on a full dataset, and acquire promising results in only about 200 seconds. This verifies the effectiveness and potential application of the model in big-data circumstances. On top of that, the testing time of AE-1SVM is a notable improvement over other methods, especially Isolation Forest and conventional OC-SVM, suggesting its feasibility in real-time environments.

\subsection{Gradient-based explanation in image datasets}
\begin{figure}[h]
\centering
\includegraphics[width=0.76\linewidth]{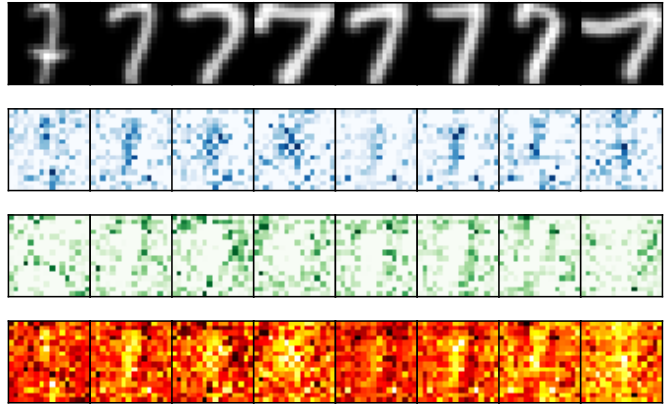}
\caption{Examples of images and their corresponding gradient maps of digits '7' in the USPS experiment. From top to bottom rows: original image, positive gradient map, negative gradient map, and full gradient map. \label{usps_gradient}}
\end{figure}

\begin{figure}[h]
\centering
\includegraphics[width=0.76\linewidth]{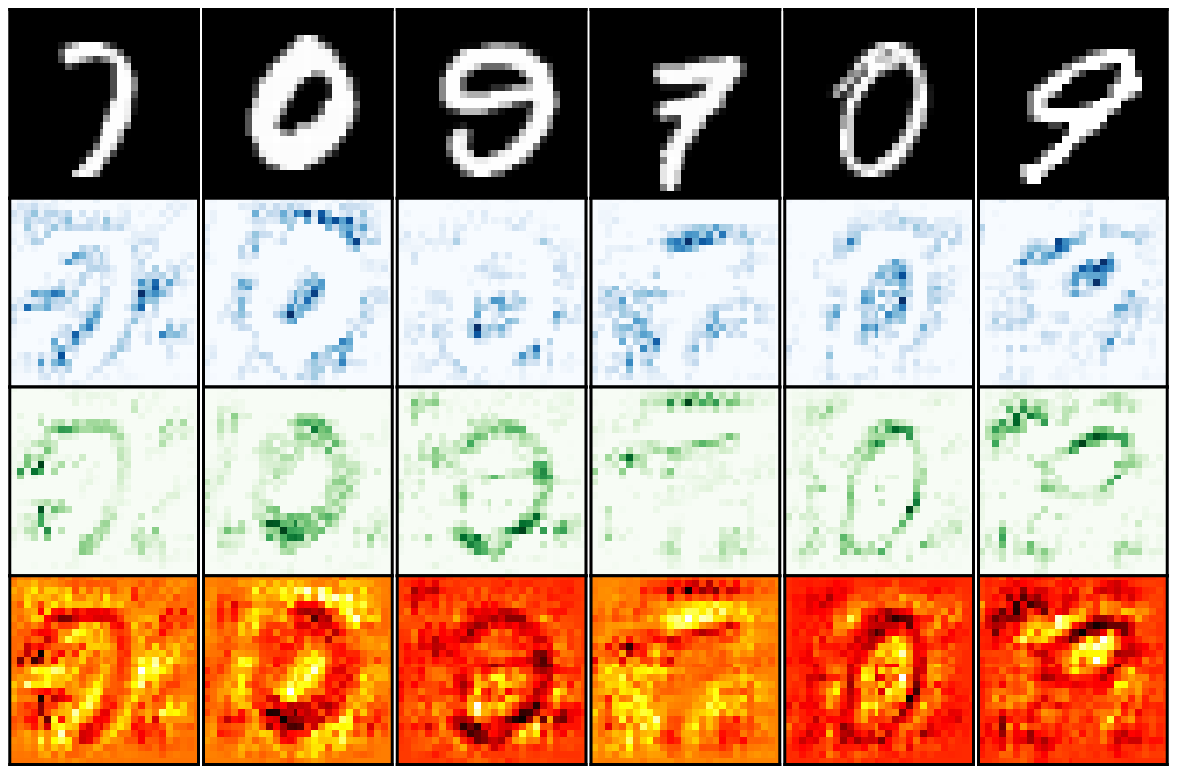}
\caption{Examples of images and their corresponding gradient maps of digits '0', '7', '9' in the MNIST experiment with convolutional autoencoder. From top to bottom rows: original image, positive gradient map, negative gradient map, and full gradient map. \label{mnist_gradient}}
\end{figure}

We also investigate the use of gradient-based explanation methods on the image datasets. Figure \ref{usps_gradient} and Figure \ref{mnist_gradient} illustrate the unsigned gradient maps of several anomalous digits in the USPS and MNIST datasets, respectively. The MNIST results are given by the version with convolutional autoencoder. 

Interesting patterns proving the correctness of gradient-based explanation approach can be observed from the Figure \ref{usps_gradient}. The positive gradient maps revolve around the middle part of the images where the pixels in the normal class of digits '1' are normally bright (higher values), indicating the absence of those pixels contributes significantly to the reasoning that the samples '7' are detected as outliers. Likewise, the negative gradient maps are more intense on the pixels matching the bright pixels outside the center area of its corresponding image, meaning that the values of those pixels in the original image exceeds the range of the normal class, which is around the zero (black) level. Similar perception can be acquired from Figure \ref{mnist_gradient}, as it shows the difference between each samples of digits '0', '7', and '9', to digit '4'.

\section{Conclusion \label{conclusions}}
In this paper, we propose the end-to-end autoencoding One-class Support Vector Machine (AE-1SVM) model comprising of a deep autoencoder for dimensionality reduction and a variant structure of OC-SVM using random Fourier features for anomaly detection. The model is jointly trained using SGD with a combined loss function to both lessen the complexity of solving support vector problems and force dimensionality reduction to learn better representation that is beneficial for the anomaly detecting task. We also investigate the application of applying gradient-based explanation methods to interpret the decision making of the proposed model, which is not feasible for most of the other anomaly detection algorithms. Extensive experiments have been conducted to verify the strengths of our approach. The results have demonstrated that AE-1SVM can be effective in detecting anomalies, while significantly enhance both training and response time for high-dimensional and large-scale data. Empirical evidence of interpreting the predictions of AE-1SVM using gradient-based methods has also been presented using illustrative examples and handwritten image datasets.

\bibliographystyle{spmpsci}
\bibliography{reference}

\end{document}